\documentclass[letterpaper, 10 pt, conference]{ieeeconf}  % Comment this line out if you need a4paper
\IEEEoverridecommandlockouts                              % This command is only needed if 
\usepackage{cite}
\usepackage{amsmath,amssymb,amsfonts}
\usepackage{graphicx}
\usepackage{textcomp}
\usepackage{xcolor}
\usepackage{tabularx}
\usepackage{multirow}

\usepackage[font=footnotesize]{caption}
\usepackage[font=footnotesize]{subcaption}

\title{Learning-Based Error Detection System for Advanced Vehicle Instrument Cluster Rendering}

\author{Cornelius Bürkle$^{1}$, Fabian Oboril$^{1}$ and Kay-Ulrich Scholl$^{1}$
\thanks{$^{1}$Intel Deutschland GmbH}}

\begin{document}

\bstctlcite{IEEEexample:BSTcontrol}

\maketitle

\thispagestyle{empty}
\pagestyle{empty}

\begin{abstract}
The automotive industry is currently expanding digital display options with every new model that comes onto the market. This entails not just an expansion in dimensions, resolution, and customization choices, but also the capability to employ novel display effects like overlays while assembling the content of the display cluster. Unfortunately, this raises the need for appropriate monitoring systems that can detect rendering errors and apply appropriate countermeasures when required. Classical solutions such as Cyclic Redundancy Checks (CRC) will soon be no longer viable as any sort of alpha blending, warping of scaling of content can cause unwanted CRC violations. Therefore, we propose a novel monitoring approach to verify correctness of displayed content using telltales (e.g. warning signs) as example. It uses a learning-based approach to separate ``good'' telltales, i.e. those that a human driver will understand correctly, and ``corrupted'' telltales, i.e. those that will not be visible or perceived correctly. As a result, it possesses inherent resilience against individual pixel errors and implicitly supports changing backgrounds, overlay or scaling effects. This is underlined by our experimental study where all ``corrupted'' test patterns were correctly classified, while no false alarms were triggered.
\end{abstract}

\section{Introduction}
Modern vehicle instrument clusters are becoming increasingly complex and are able to show an ever increasing amount of information paired with a growing degree of user design configurations. Nowadays, all data including the speedometer, telltales (signal lights) and navigation are digitally rendered and simple analog designs belong to the past~\cite{vasantharaj2014state,lyu2022study}. The advantage of the digital displays is obvious, as these can be customized and updated easily, and additionally enable modern display effects such as enlargements (scaling), warping or overlay of information on top of diverse backgrounds. While this new freedom offers a considerable level of design flexibility, it also raises the challenge to ensure that the rendered content is correct, and that rendering/composition errors often need to be detected immediately to apply appropriate countermeasures. For example, an error in the navigation part is certainly inconvenient, and even worse is a corrupted telltale for a failed brake system. Hence, it is important that such instrument clusters are paired with adequate error detection systems to ensure that all rendered content is correct~\cite{anisimov2023smartcluster,vesa2023standard,tiClusterSafety}.

\begin{figure}[t!]
    \centering
    \includegraphics[width=0.7\columnwidth]{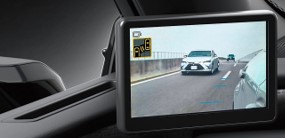}
    \caption{Example display using an overlay of a telltale on a camera image~\cite{lexusMirror}. Classical error detection methods are hardly applicable for these emerging rendering effects. This paper proposes a novel and very effective solution.}
    \label{fig:displayclusterexample}
    \vspace{-0.4cm}
\end{figure}

The standard approach to ensure correctness of displayed information is to use classic error detection codes such as CRC to verify the video/image stream after composition but before it is sent to the display matrix. However, these codes are approaching their limits, in a world in which automakers started to use overlay effects (see Figure~\ref{fig:displayclusterexample}) and in a near future in which users will be able to customize the look of the display cluster, the background, the rendering effects, the colors or the placement of information. For example, if an icon color is modified from red to orange due to an overlay effect with a greenish background, CRC may flag this modification as erroneous if the CRC check value is not updated as well. This however is not trivial, requires expensive hardware modifications, and may not be possible for all rendering effects~\cite{canto2020reliable}. \textit{Hence, new more flexible error detection approaches are required for advanced digital instrument clusters.} 

Therefore, we propose a novel learning-based error detection system for the complex rendering process of modern vehicle instrument clusters using telltales as example. Our system uses an anomaly detection approach which analyzes rendered telltales to detect errors that would cause a wrong human perception of the telltale. A major advantage of this system over state-of-the-art is generalization combined with an inherent robustness against minor disturbances, i.e. rendering errors that do not alter the human understanding such as a missing pixel do not cause an exception, and advanced rendering effects such as alpha blending are supported as well. Our results show that all ``corrupted'' test samples, i.e. those that were not clearly perceivable by a human, were properly classified as erroneous, while no false alarms were found in fault-free scenarios.

The remainder of the paper is organized as follows. In Section~\ref{sec:background} the problem of rendering errors in digital instrument clusters is introduced. Afterwards, in Section~\ref{sec:solution} our novel solution is presented, followed by an experimental evaluation in Section~\ref{sec:results}. Finally, Section~\ref{sec:conclusion} concludes the paper.

\section{Background}
\label{sec:background}

\begin{figure}[t!]
    \centering
    \includegraphics[width=0.8\columnwidth]{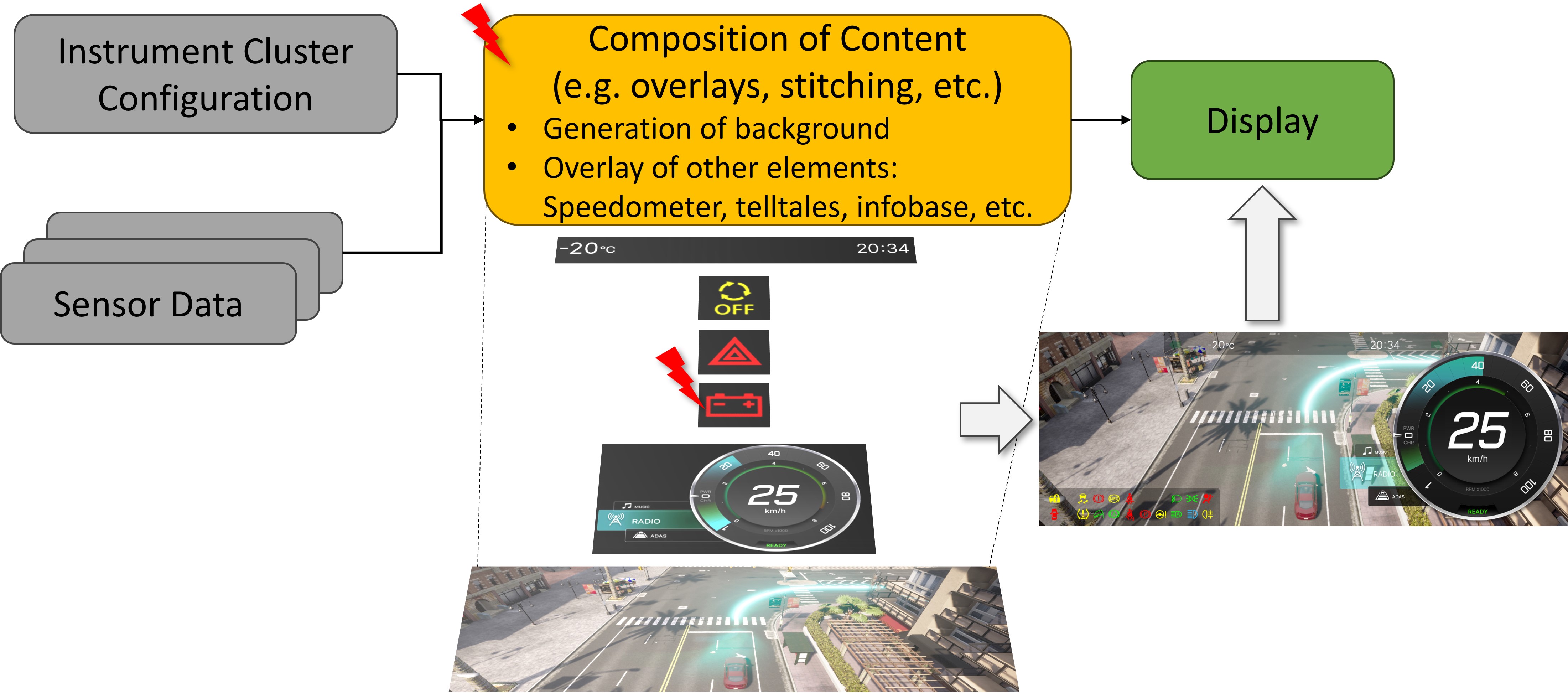}
    \caption{Illustration of display composition including some example telltales, navigation and speedometer. Flash indicates possible error source.}
    \label{fig:data_flow}
    \vspace{-0.4cm}
\end{figure}

Modern vehicle instrument clusters use a variety of data sources that need to be aggregated into one image (or video stream) that can then be sent to the display for visualization (see Figure~\ref{fig:data_flow}). This composition involves many processing steps at various levels like converting CAN messages with vehicle status information to telltales which then need to be stitched and overlayed in multiple steps. Consequently, various errors can affect the finally displayed image, which can occur at any of the composition stages, as described in~\cite{axmann2020supervising}. It is important to note that some errors can impair only one composition element, for example a single telltale, while the rest of the image is perfectly rendered. Errors in the finally displayed image may include corruption or artifacts (e.g. wrong pixels), distortion (wrong shape), erroneous zoom factor (wrong size), erroneous orientation/color/contrast or brightness to name only the most relevant types~\cite{axmann2020supervising,bauer2019neue}.

\begin{figure}[b!]
    \centering
    \vspace{-0.2cm}
    \includegraphics[width=0.9\columnwidth]{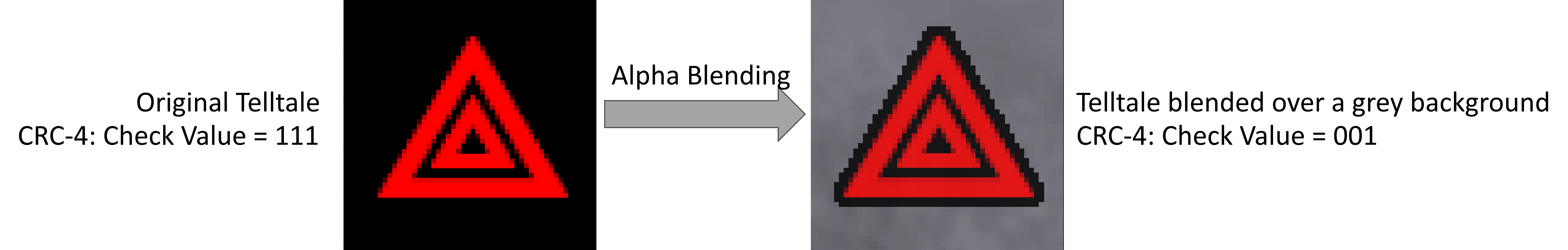}
    \caption{Twice the same telltale, but on the right laid over a grey background result in different CRC check values}
    \label{fig:telltale_crc}
\end{figure}

An important aspect of the content composition is that it can involve non-linear image processing functions, i.e. $I_{out} = f(I_{in})$, where $f$ is not a linear function. In the past, composition was often very simple as elements where usually placed next to each other without being modified or rendered on top of each other using e.g. blending effects. Hence, CRC checks were an effective mean to detect any error during the execution of the composition function $f$~\cite{vesa2023standard}. However, when for example alpha blending is used, the CRC check value can change, as illustrated in Figure~\ref{fig:telltale_crc}. This requires either expensive hardware modifications to calculate CRC for each render step (e.g.~\cite{canto2020reliable}) or other approaches to detect errors due to rendering faults.

In this regard it is worth noting that rendering/composition errors leading to the display of information that a human cannot perceive correctly can be of mixed criticality. If the error affects only a convenience function, e.g. a part of the navigation map is not displayed correctly, it is surely undesired yet tolerable. If in contrast, a warning light about a failed brake system is not correctly perceived, it is a safety critical event. Thus, error detection solutions are required, that can handle both categories~\cite{clusterSafetyWhitePaper}.

\begin{figure}[t!]
    \centering
    \includegraphics[width=0.95\columnwidth]{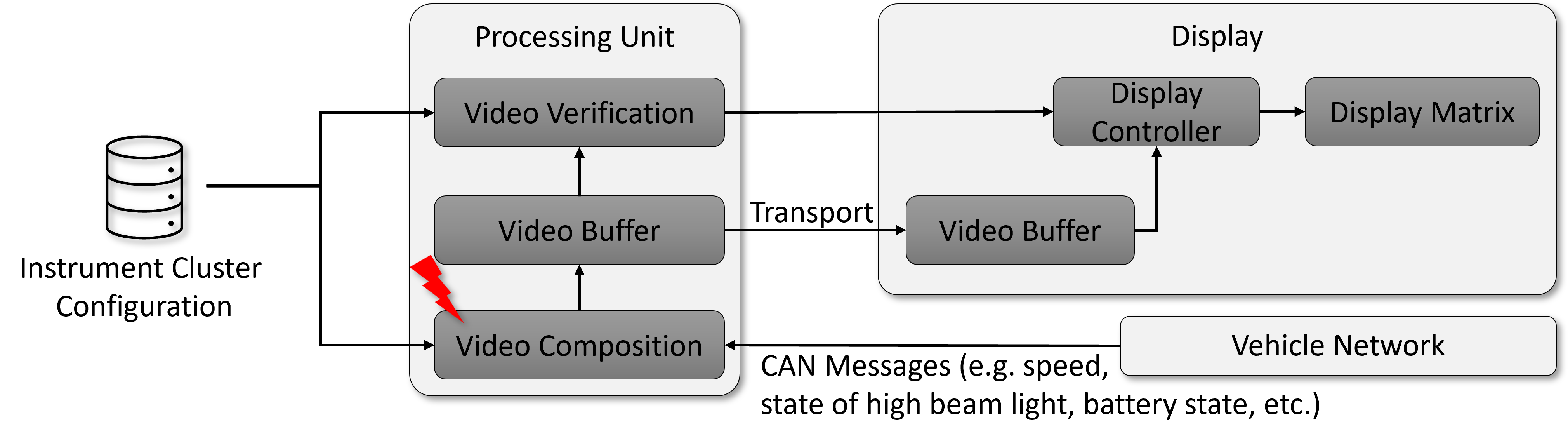}
    \caption{Components involved to compose, verify and visualize data on a display. In this work we address errors within the video composition step.}
    \label{fig:pipeline}
    \vspace{-0.4cm}
\end{figure}

To address the problem of rendering/composition errors,~\cite{anisimov2023smartcluster} proposed an architectural solution depicted in Figure~\ref{fig:pipeline}, which uses additional checks to verify the video/image stream after composition but before it is sent to the display matrix. An abstract description of such an error detector, also referred to as ``monitor'' or ``checker'' is provided in~\cite{gulati2018data}, which proposed to use a software-based calculation of the check values to be compared including CRC, MISR (multiple independent signature registers) or hash values. While this enables the handling of alpha blending or scaling, single pixel errors or other non-perceivable deviations will be treated as errors. An alternative is to use abstract descriptors to detect errors, as suggested by~\cite{conrad2021improving}.

Another possibility to validate the final display output is to use diodes~\cite{conrad2021improving} or even a camera-based monitor~\cite{blankenbach2021advanced}. However, these come at additional costs and limited flexibility, which seems out-of-scope for most automotive applications. 

Finally, it is worth mentioning that AI-based error detection approaches have been proposed. For example,~\cite{willers2020safety,buerkle2020onlineenv} presented monitor concepts for perception systems that partially uses AI. In summary, AI can be used under certain conditions, including a representative training and validation set, and the use of out-of-distribution monitoring. 

In conclusion, a new composition/rendering error detection approach for advanced digital instrument clusters is required. The solution needs to be robust and effective, support high resolution displays, user customizable rendering effects and is robust against errors that humans will not realize or can tolerate. In this paper, we propose a novel learning-based approach that fulfills all these characteristics. Is uses an anomaly detection concept on abstract features  to verify the correctness, which is explained in depth in the next section.

\section{Proposed Telltale Error Detection System}
\label{sec:solution}

In this section, we will introduce our novel learning-based error detection approach for rendering using telltales as example. The goal is to realize a so-called \textit{telltale monitor} that verifies the rendered output on-the-fly using abstract features rather than pixel-level data, can handle overlays (alpha blending) of multiple telltales on varying backgrounds, effects such as warping or scaling and is resilient against a small number of pixel errors.

Please note that after an error is detected, various means can be employed to mitigate the effects, depending on the criticality of the event (see Section~\ref{sec:background}). In case of the telltales, it may be desired to provide a warning message and possibly re-render a default icon at a replacement location on a fixed (known) background which can be checked with classical means (e.g. CRC). As this is rather straightforward, we focus in this work on the error detection capabilities.

\subsection{High-Level Overview}

\begin{figure}[t!]
    \centering    
    \includegraphics[width=0.95\columnwidth]{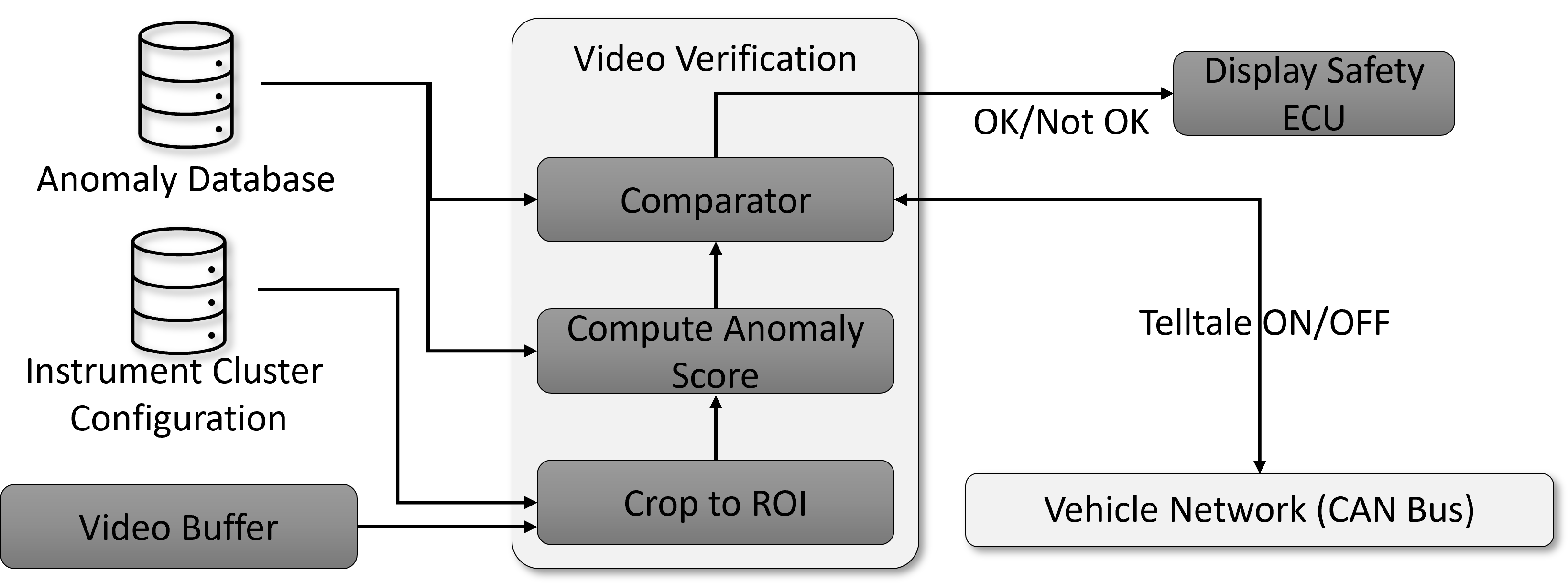}
    \caption{Components of our proposed telltale monitor.}
    \label{fig:normal_operation_when_telltale_on}    
    \vspace{-0.4cm}
\end{figure}

Our proposed telltale monitor, depicted in Figure~\ref{fig:normal_operation_when_telltale_on}, is designed to be permanently active to detect rendering errors that affect the visibility of an active telltale as well as faults that lead to the display of a telltale that should be inactive. The monitor treats the entire rendering step as a black box and only checks if the final result contains any perceivable errors. Therefore, the monitor is realized as a software module comprising multiple processing steps to verify the correct composition of the cluster content before it is handed over to the display.

The first step is to retrieve the required video/image data from the video buffer. Additionally, the location of the telltale is obtained from the instrument cluster configuration storage (e.g. an EPROM). With this input data, the image, which can be quite large (e.g. 4K resolution), is cropped to a smaller region of interest (ROI) around the telltale.

This smaller image (e.g. 52x52 pixels), is then fed into the core component of the system to compute the \textit{anomaly score}. This step is based on the work from~\cite{ndiour2022fre}, and will be described in detail in Section~\ref{sec:anomaly}.

Finally, the resulting anomaly score is compared against a pre-defined threshold, which can depend on whether the telltale should be displayed or not. The comparison success is dependent on the telltale state, i.e. if the telltale should be ON, scores below the threshold represent a successful verification and can be displayed, while all other results\footnote{Some telltale examples are provided in Figure~\ref{fig:errors_examples}} invoke an error handling routing by the display ECU. However, if the telltale should not be displayed, low scores indicate that the telltale is erroneously part of the image. Hence, these are classified as failure, while scores above the threshold represent a success in this situation. The result ('OK' or 'NOK') is then sent to the display ECU.

Please note in case that multiple telltales should be displayed at the same time, the system will create multiple crops for the different ROIs and then perform all the required following steps in parallel for all telltales.

\subsection{Anomaly Detection Step}
\label{sec:anomaly}

\begin{figure}[t!]
    \centering
    \includegraphics[width=0.95\columnwidth]{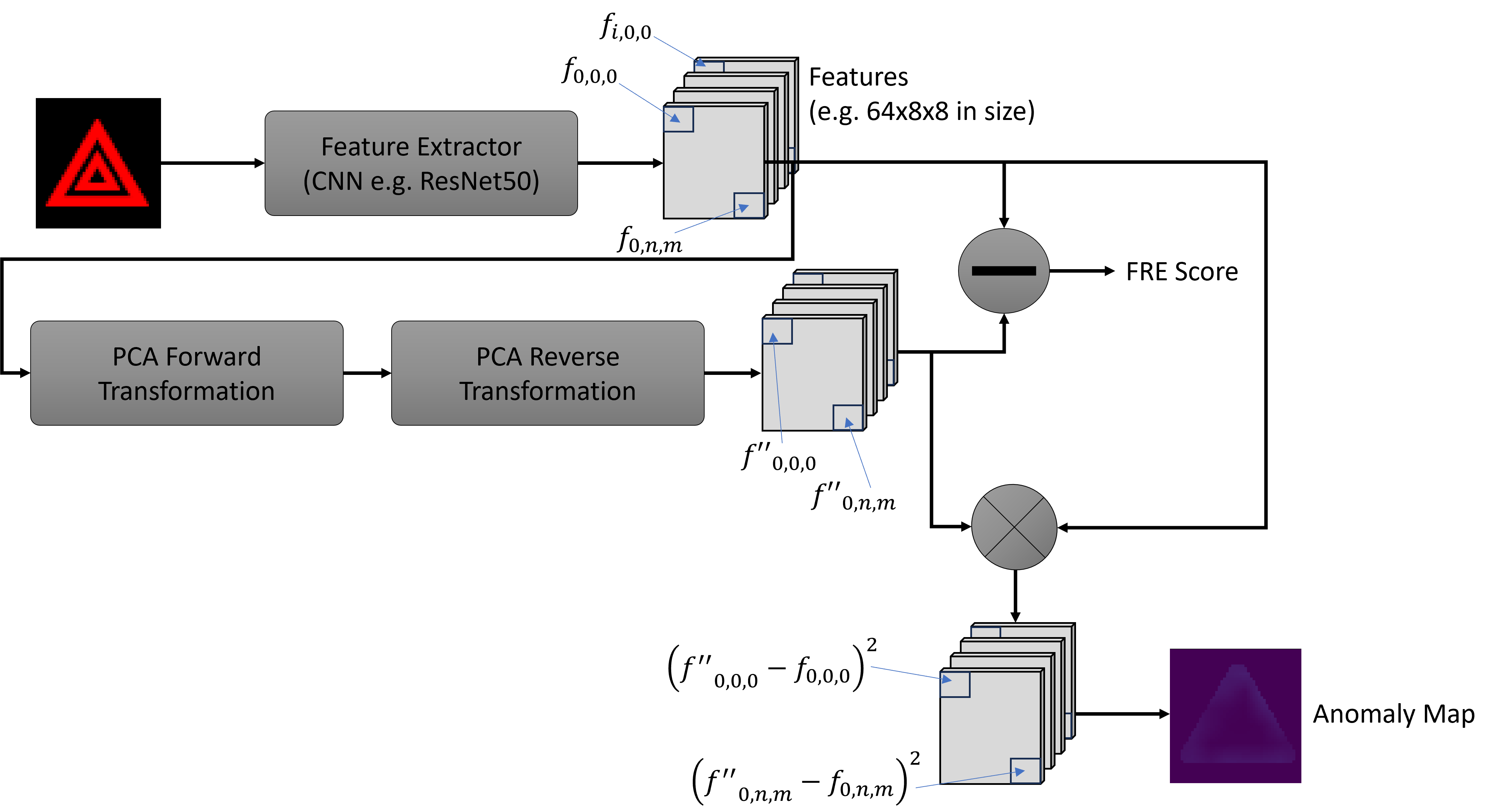}
    \caption{Anomaly scoring approach (based on~\cite{ndiour2022fre})}
    \label{fig:PCA}
    \vspace{-0.5cm}
\end{figure}

The core element of our telltale monitor, is the rendering anomaly detection step. For this component, we make use of the anomaly detection system proposed in~\cite{ndiour2022fre}. The idea of this system is to use an input image, feed it into a feature encoder of a CNN\footnote{In this work we use a pre-trained ResNet-18~\cite{he2016deep} as CNN.}, and then use a principal component analysis (PCA) on the feature map to analyze if the input image contains any kind of anomaly, in our case any kind of error/corruption. Therefore, the system calculates the feature reconstruction error (FRE), as depicted in Figure~\ref{fig:PCA}. For this purpose, the feature tensor $f$ from an intermediate CNN layer, for example belonging to the first or second layer, is extracted\footnote{In this work we use the first layer.}. Next, $f$, which is originally in a high-dimensional space $\mathcal{F}$, is transformed with the help of a PCA, i.e a transformation $\mathcal{T} : \mathcal{F} \rightarrow \mathcal{S}$, with $\text{dim}(\mathcal{F}) >> \text{dim}(\mathcal{S})$ is applied on $f$. The result is $f' = \mathcal{T}(f)$. Afterwards, the inverse transformation $\mathcal{T}^{-1}$ is applied on $f'$ resulting in:
\begin{equation}
    f'' = \mathcal{T}^{-1}(f') = \mathcal{T}^{-1}(\mathcal{T}(f))
\end{equation}
Finally, the FRE is calculated as follows:
\begin{equation}
   \text{FRE}(f) = || f'' - f ||
\end{equation}

The goal is that valid images achieve a near-zero FRE score. Therefore, during the PCA fitting process only valid telltales without errors are provided to derive the transformation function $\mathcal{T}$. This can be interpreted as finding a multi-dimensional ellipsoid that encloses the most relevant feature points. At runtime, any valid telltale image that is given to the transformation process, will only use features which are already part of the PCA, i.e. that falls within the ellipsoid. Thus the forward and inverse transformations will cancel each other. Consequently, the FRE in such a cases will be close to zero according to Equation (2). However, if an error (anomaly) is introduced in the telltale, the transformation steps will lose part of the information contained in the ``erroneous'' tensor $f$ (those that are outside of the ellipsoid), i.e. $f'' \neq f$ and as a result $\text{FRE} >> 0$.

 Under normal operating conditions, with changing background images (for example due to a running navigation with changing map content), the ideal case with $\text{FRE} = 0$ will usually not occur. However, the scores for telltales with visible errors and significant degradation will be considerably different from scores of acceptable telltales. Hence, a threshold $\tau$ can be introduced to differentiate among acceptable telltales (even if these include a few non noticeable faults), and telltales that are clearly corrupted with the risk that a human driver may miss-perceive the telltale. Based on this, we can define the set of acceptable ('OK') telltales as the set of all telltales with $\text{FRE} < \tau$.
 \begin{equation}
     \mathcal{A} = \left\lbrace \text{telltale image} ~|~ \text{FRE}(f) < \tau \right\rbrace
 \end{equation}

It is worth noting that the set of fault free telltales, which is used to train the PCA, is a true subset of $\mathcal{A}$, and that any telltale that is not within $\mathcal{A}$, i.e. with $\text{FRE} \geq \tau$ is classified as erroneous ('NOK' for 'not OK'). This comparison step is the last component illustrated in Figure~\ref{fig:normal_operation_when_telltale_on}, and if it is above the threshold may trigger an error mitigation action.

Both $\tau$ as well as the size of the training set to derive a meaningful representation of $\mathcal{T}$ need to be carefully chosen. If $\tau$ is too small, it may cause flagging valid telltales as erroneous, which can be undesired. Similarly, if $\tau$ is too large, telltales with visibly errors could be classified as 'OK'. Also, if the training set it too small, not all relevant features may be represented well enough by $\mathcal{T}$, resulting in the same effect as a $\tau$ that is too small. However, as we will show in Section~\ref{sec:results} a few hundred training and testing samples are sufficient to obtain a meaningful transformation function $\mathcal{T}$ and definition of $\tau$.

Finally, it is worth noting that it is not required to train feature extractor itself on a telltale-specific dataset. Instead, any abstract feature representation can be used as input to the PCA. For this reason, we used a standard pre-trained ResNet-18 in this work as feature extractor. In fact, we use the first feature layer of ResNet-18, which consists of 64 different convolutions, each with a size of 8x8, 16x16 or 32x32 depending on the original image size.~\cite{ndiour2022fre} suggests to fit a single PCA covering all convolutions, which however has certain drawbacks. A major one being the size of the resulting transformation matrix (e.g. 65536x65536), that requires a lot of memory and computational resources, which both are typically rare for a SM. Therefore, we analyzed the impact of each convolution and identified that only a few are really of importance. Consequently, we propose to fit a PCA for each individual convolution that is of relevance. As explained in Section~\ref{sec:results}, this helps to improve the robustness as well as the computational demands.

\begin{figure}[b!]
    \centering    
    \includegraphics[width=0.95\columnwidth]{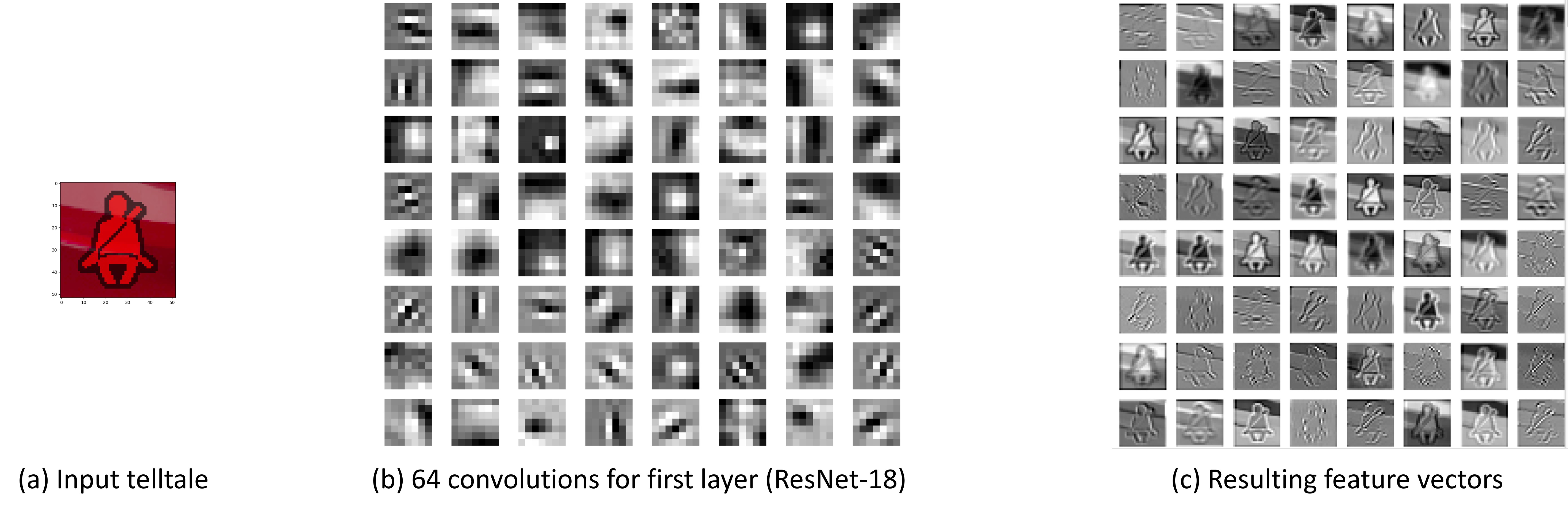}
    \caption{Visualization of the different feature vectors of a given telltale}
    \label{fig:Convolutions}    
\end{figure}

\subsection{FRE Score Hardening}
As we discussed before, the decision if a telltale is 'OK' or 'NOK' is based on a single value comparison, which of course by the nature of the employed dimension reduction (within PCA) can introduce undesired effects. One of these being a false classification either being 'NOK' instead of 'OK' or vice versa. To mitigate these, we propose an enhanced FRE scoring.

\begin{figure}[t!]
    \centering
    \includegraphics[width=0.75\columnwidth]{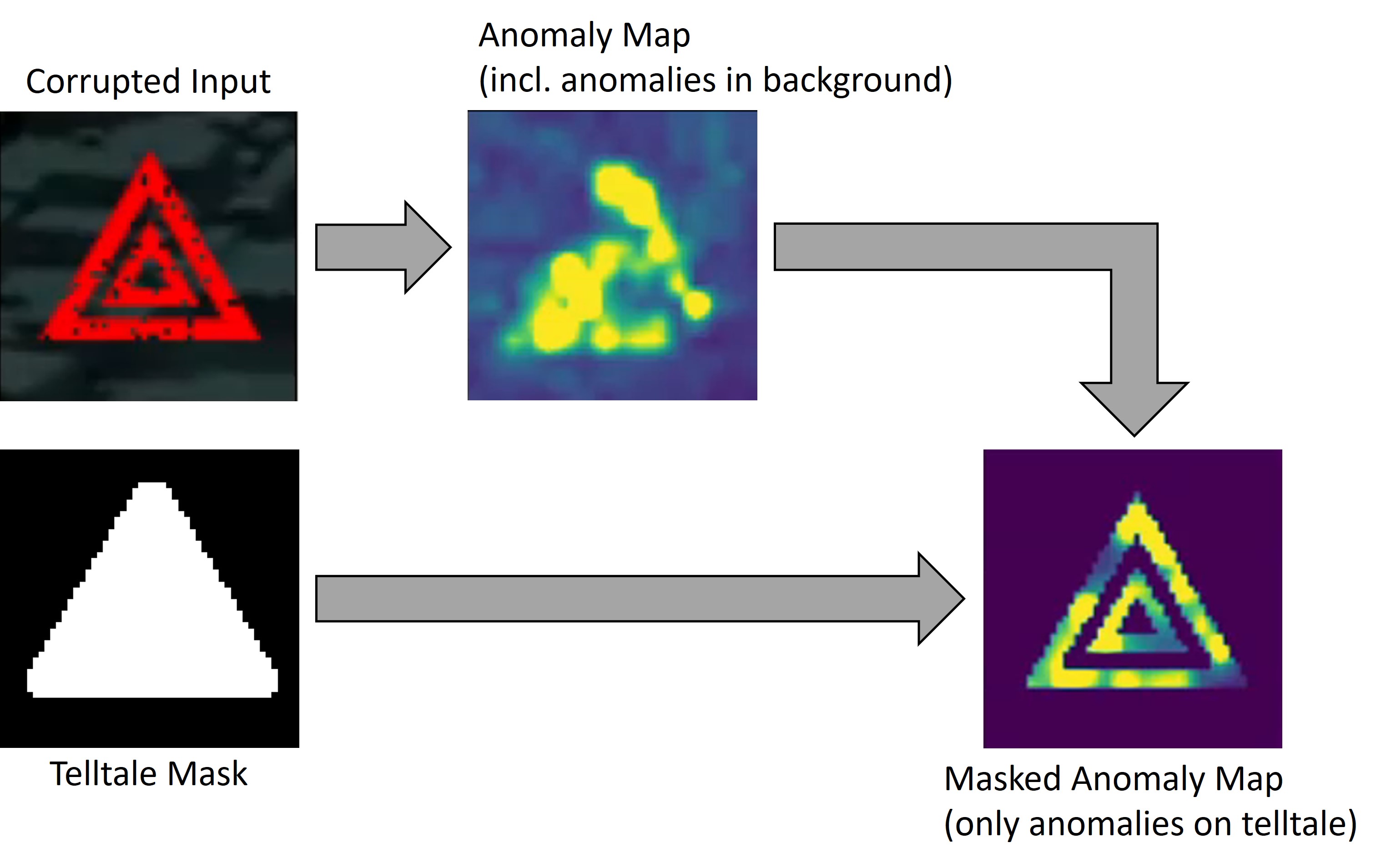}
    %\vspace{-0.1cm}
    \caption{Left: Rendered telltale with errors (missing pixels). Middle: Anomaly map considering all pixels. Right: Anomaly map masked by the telltale shape.}
    \label{fig:masked_score}
    \vspace{-0.4cm}
\end{figure}

Therefore, we use the \textit{anomaly map} or FRE map. This map can be interpreted as an image, in which a pixel $i$ reflects the difference $(f''_i - f_i)^2$, as illustrated in Figure~\ref{fig:PCA}. Visually this map can be interpreted as a differential image, i.e. the brighter an image, the greater is the anomaly (error) in this region. This is also shown by some examples in Figure~\ref{fig:errors_examples}. 

Now, instead of considering all pixels (or elements of the tensors $f$ and $f''$) as in Equation~(2), one can also restrict the analysis to those pixels that are of interest, i.e. those pixels that are part of the telltale's shape but ignore pixels that are part of the background. To achieve this, a mask can be generated using the shape of the telltale  and subsequently applied to the anomaly map, as depicted in Figure~\ref{fig:masked_score}. The result is a score that is only influenced by anomalies (or faults) on the telltale, which is more robust than a score over the entire map. Of course, in a similar way also a weighted score is possible by applying different weights to background pixels and pixels belonging to the telltale shape.

Another improvement addresses the robustness of the FRE calculation itself. Following Equation~(2), the FRE score can be dominated by a single tensor element, for example if most of the elements of $f$ and $f''$ are similar except one. If for this element $i$ the difference $||f''_i - f_i||$ is much larger than for any other element, it can hold $\text{FRE}(f) \approx ||f''_i - f_i||$. This is of course undesired, as computational errors causing only a single bit (or few bits) to be considerably off (e.g. by affecting an exponent bit), would influence the entire result. Therefore, we propose to limit the values within the FRE calculation to be no larger than a predefined value $d_{max}$. 

In summary, Equation~(2) can be reformulated using the Euclidean norm (of course also other distance measures could be used) to:
\begin{equation}
   \text{FRE}(f) = \frac{1}{|\mathcal{P}|}\sqrt{\sum_{i\in\mathcal{P}}min\left\lbrace {(f''_i-f_i)^2,d_{max}} \right\rbrace},
\end{equation}
where $\mathcal{P}$ is the set of relevant pixels (e.g. just on mask). 

Finally, it is important to note that the decision 'OK'/'NOK' may not be taken on the result of a single image (frame) only. Instead the FRE scores over multiple frames can be combined. By this means, short upsets that only occur in very few frames do not disturb the final signal. Instead these can be filtered out, making the system more robust. One possible filtering approach is to use a running average over the FRE scores over a sliding time window, e.g. last 60 frames (approx. 1 second in good displays). Hence, a false score in a few frames will not considerably alter the running average, while a permanent issues will become visible within 1 second. Similarly, if the error disappears, the system will switch to 'OK' within the next second.

\subsection{Testing Mode}
So far, we discussed the normal operation of the telltale monitor. If a telltale is inactive, we foresee an additional mode that can be used to test if everything works as expected.

In this mode, instead of fetching the image from the video buffer (see Figure~\ref{fig:normal_operation_when_telltale_on}), a test image from a database is fed into the telltale monitor, for which the anomaly map is computed. The result is then compared to a reference map from the database for the given test image. If any difference at pixel-level is detected, it indicates a problem with the anomaly scoring stage and an error mitigation action can be initiated. In addition the anomaly scoring stage can be reset (i.e. refresh all memories, re-run steps, check if problem still exists), to detect if this was a single event upset or a more persistent issue that the driver should be informed about.

Please note that both good as well as corrupted test images can be used during test mode to verify both, correct detection of 'OK' as well as of 'NOK' telltales. Furthermore, it is worth noting that the monitor in testing mode is not connected to the rest of the display and rendering pipeline, meaning the visual output of the instrument cluster is not altered such that this process remains hidden from the driver.
\section{Experimental Evaluation}
\label{sec:results}

\subsection{Setup, Dataset Choices \& Rendering Errors }

\begin{figure}[b!]
     \centering
     \vspace{-0.6cm}
     \begin{subfigure}[b]{0.25\columnwidth}
         \centering
         \includegraphics[clip,trim=0.0cm 0.cm 0.0cm 0.cm,width=0.35\textwidth]{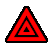}
         %\vspace{-0.2cm}
         \caption{ Warning }
         \label{fig:subFigWarn}
     \end{subfigure}
     %\hfill
       \begin{subfigure}[b]{0.25\columnwidth}
         \centering
         \includegraphics[clip,trim=0.0cm 0.cm 0.0cm 0.cm,width=0.35\textwidth]{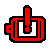}
         %\vspace{-0.2cm}
         \caption{ Engine }
         \label{fig:subFigEngine}
     \end{subfigure}
       % \hfill
       \begin{subfigure}[b]{0.25\columnwidth}
         \centering
         \includegraphics[clip,trim=0.0cm 0.cm 0.0cm 0.cm,width=0.35\textwidth]{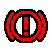}
         %\vspace{-0.2cm}
         \caption{ Brake }
         \label{fig:subFigBrake}
     \end{subfigure} 
     \\
        %\hfill
       \begin{subfigure}[b]{0.25\columnwidth}
         \centering
         \includegraphics[clip,trim=0.0cm 0.cm 0.0cm 0.cm,width=0.35\textwidth]{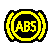}
         %\vspace{-0.2cm}
         \caption{ ABS }
         \label{fig:subFigABS}
     \end{subfigure}
        %\hfill
       \begin{subfigure}[b]{0.25\columnwidth}
         \centering
         \includegraphics[clip,trim=0.0cm 0.cm 0.0cm 0.cm,width=0.35\textwidth]{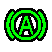}
         %\vspace{-0.2cm}
         \caption{Autopilot}
         \label{fig:subFigAuto}
     \end{subfigure}
        %\hfill
       \begin{subfigure}[b]{0.25\columnwidth}
         \centering
         \includegraphics[clip,trim=0.0cm 0.cm 0.0cm 0.cm,width=0.35\textwidth]{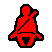}
         %\vspace{-0.2cm}
         \caption{ Seatbelt }
         \label{fig:subFigSeat}
     \end{subfigure}
        %\vspace{-0.2cm}
        \caption{Exemplary telltales used for evaluation}
        \label{fig:telltales} 
%        \vspace{-0.2cm}
\end{figure}

\begin{figure}[t!]
     \centering
     \begin{subfigure}[b]{0.48\columnwidth}
         \centering
         \includegraphics[clip,trim=0.0cm 0.cm 0.0cm 0.cm,width=\textwidth]{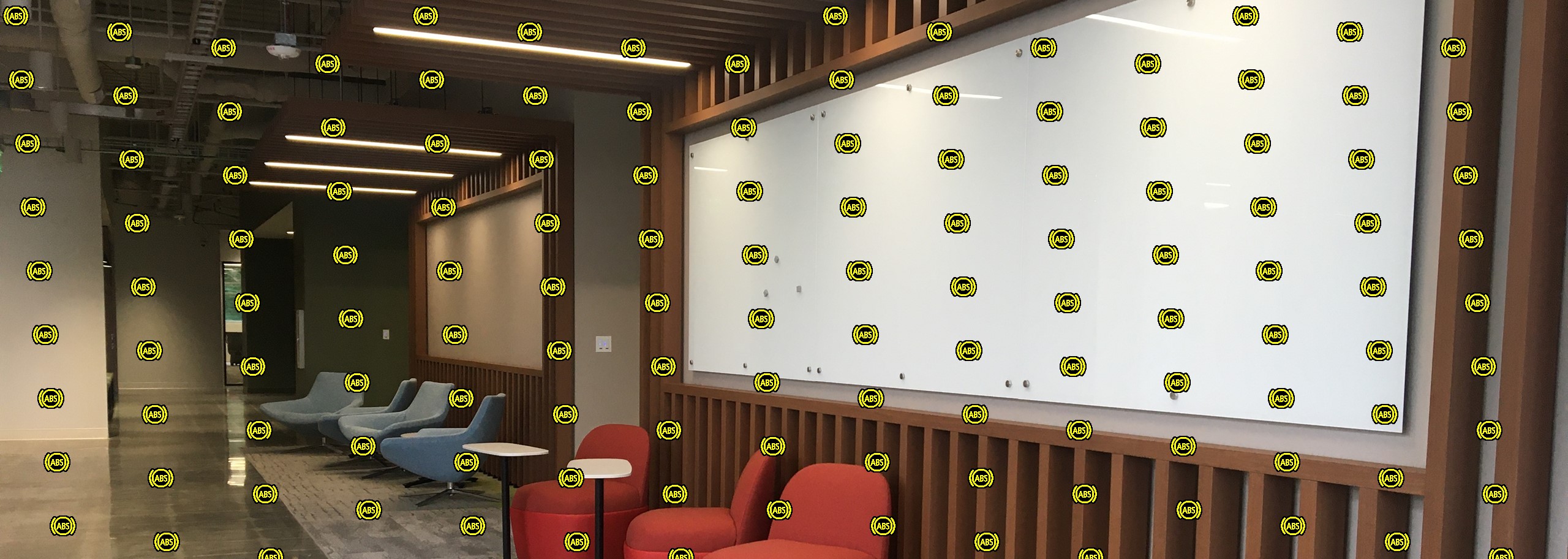}
         \caption{ random background image }
         \label{fig:subFigRandomBackground}
     \end{subfigure}
     \hfill
     \begin{subfigure}[b]{0.48\columnwidth}
         \centering
         \includegraphics[clip,trim=0.0cm 0.cm 0.0cm 0.cm,width=\textwidth]{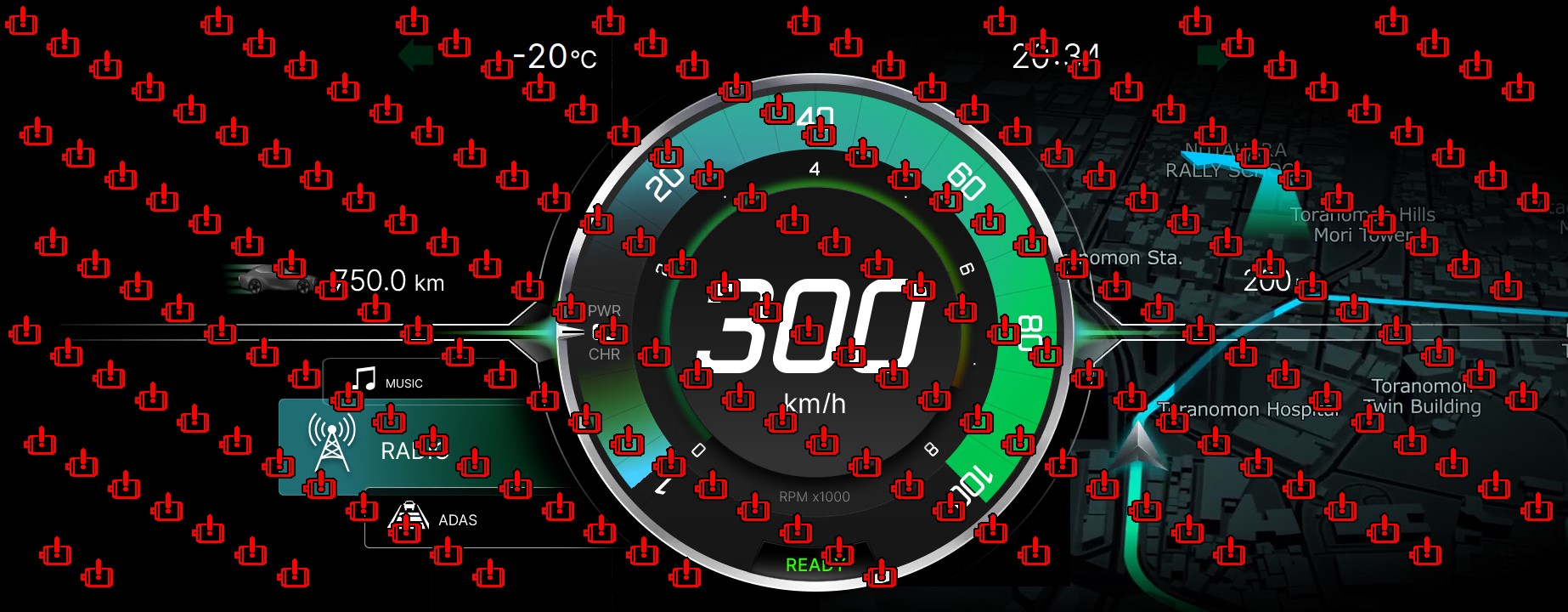}
         \caption{target domain }
         \label{fig:subFigTargetBackground}
     \end{subfigure}
        %\vspace{-0.1cm}
        \caption{Exemplary training images with overlayed telltales}
        \label{fig:training_images}
%        \vspace{-0.2cm}
\vspace{-0.4cm}
\end{figure}

\begin{figure}[b!]
    \vspace{-0.4cm}
    \centering
     \begin{subfigure}[b]{0.95\columnwidth}
    \includegraphics[clip,trim=0.0cm 0.5cm 0.0cm 0.cm,width=\textwidth]{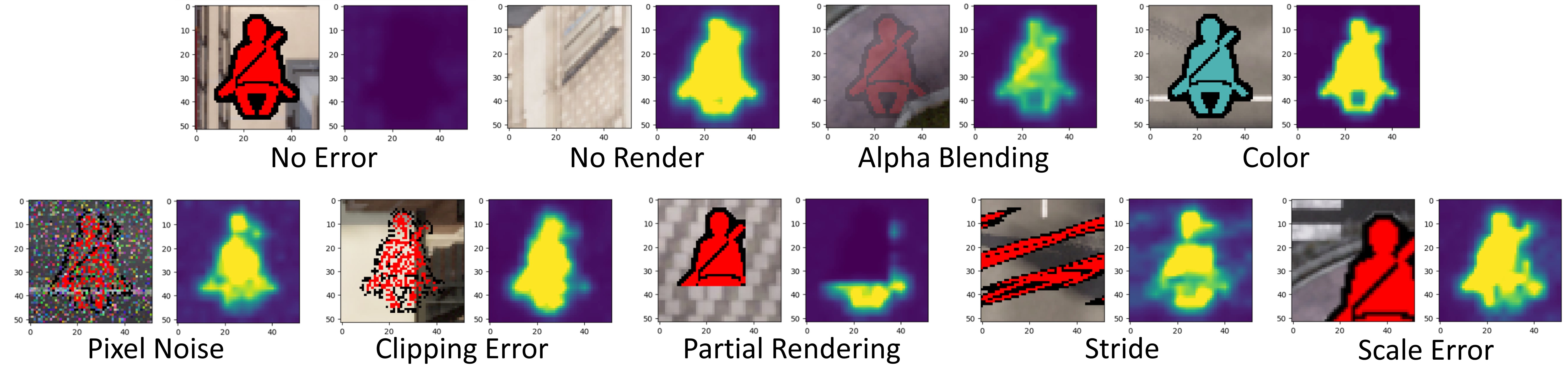}
    \caption{}
    \label{fig:errors_examples}
    \end{subfigure}
    \\
    \centering
     %\hfill
       \begin{subfigure}[b]{0.2\columnwidth}
    \includegraphics[clip,trim=0.0cm 10.cm 25.0cm 0.cm,width=\textwidth]{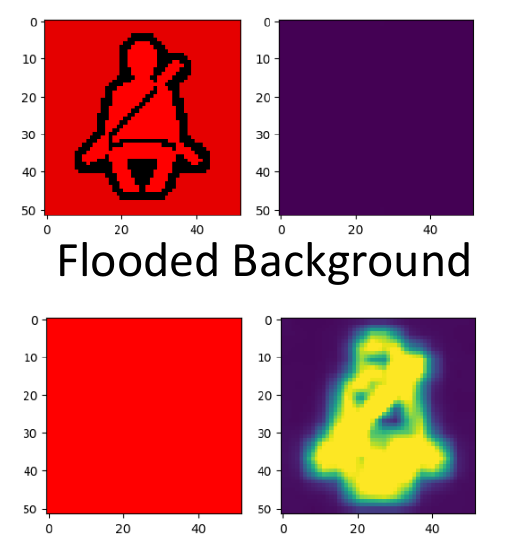}
    \vspace{-0.5cm}
    \caption{}
    \label{fig:errors_flood}
    \end{subfigure}
    
    \caption{Examples of telltales errors for evaluation with resulting anomaly maps}
    \vspace{-0.2cm}
\end{figure}

For the evaluation we tested our solution for a selection of six different telltales shown in Figure~\ref{fig:telltales}, that are based on the Automotive Grade Linux sample application\footnote{https://github.com/agl-ic-eg/cluster-refgui}. We modified these slightly, to have a black edge which favors robustness of the approach, but in general any telltale could be used.
We used ResNet-18 as feature extractor with an input images of size 128x128 pixel. This created a feature tensor of size 64x16x16. We also tested a smaller input sizes down to 64x64 pixel and larger images with 256x256 pixel, as well as a ResNet-50. In all cases, the results were basically the same, only the actual anomaly score values differed among the configurations. Thus, we will only show the results for ResNet-18 and 128x128 input images. For this configuration, the runtime on a single E-core of an Intel Core i9-13900K for a C++ implementation of the entire pipeline is around 5ms, i.e. 200 frames per second can be analyzed.

To train the PCA (see Section~\ref{sec:anomaly} for details), and evaluate the performance of our detector, we used three different datasets:
\begin{enumerate}
    \item A training dataset that only contains perfect renderings (4000 images per telltale).
    \item A test dataset that contains both corrupted as well as perfect telltales (1800 images per telltale).
    \item An evaluation dataset that is similar to the test dataset but is 11700 images large (per telltale).
\end{enumerate}

The training dataset used various background images to make the training as diverse as possible. Therefore, we placed the telltales at random positions (see Figure~\ref{fig:training_images}) and then cropped them to the expected input size. Additionally, some actual backgrounds from the target application (here: an enhanced version of the Automotive Grade Linux reference GUI~\cite{AGL}) were used. The purpose of the test dataset is to determine the threshold $\tau$ (see Equation~(3)), to separate 'OK' from 'NOK' input images. Therefore, it comprised 200 images for each error type as well as 200 perfect samples (1800 images in total). Finally, the evaluation was performed on the evaluation dataset, which contained 1300 images for each defect type.

\begin{figure}[t!]
     \centering
     \begin{subfigure}[b]{0.49\columnwidth}
     \vspace{-0.1cm}
         \centering
         \includegraphics[clip,trim=0.0cm 0.cm 0.0cm 0.cm,width=\textwidth]{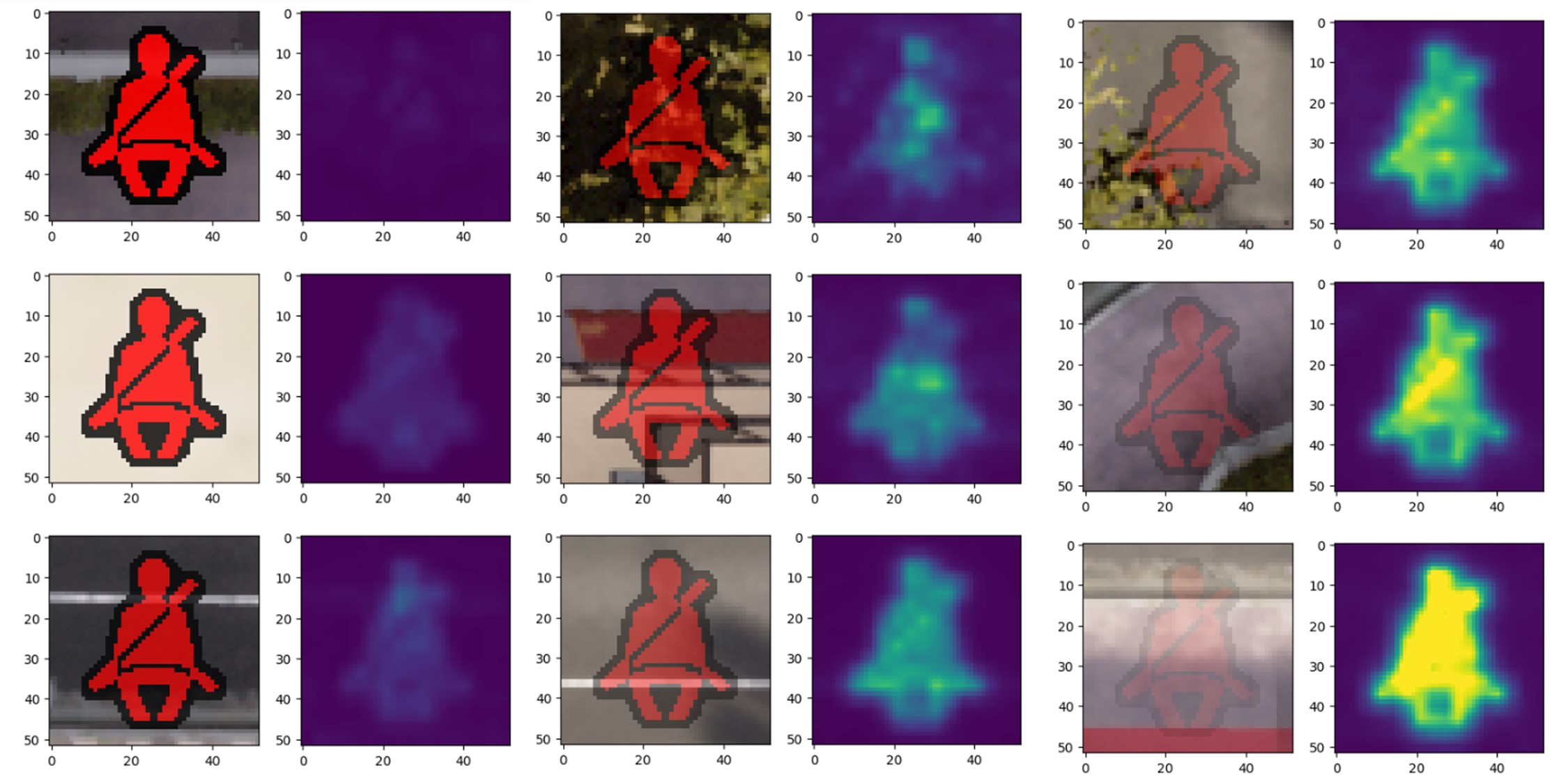}
         \caption{ Alpha Blending}
         \label{fig:subFigAlphaBlend}
     \end{subfigure}
     \hfill
     \begin{subfigure}[b]{0.49\columnwidth}
         \centering
         \includegraphics[clip,trim=0.0cm 0.cm 0.0cm 0.cm,width=\textwidth]{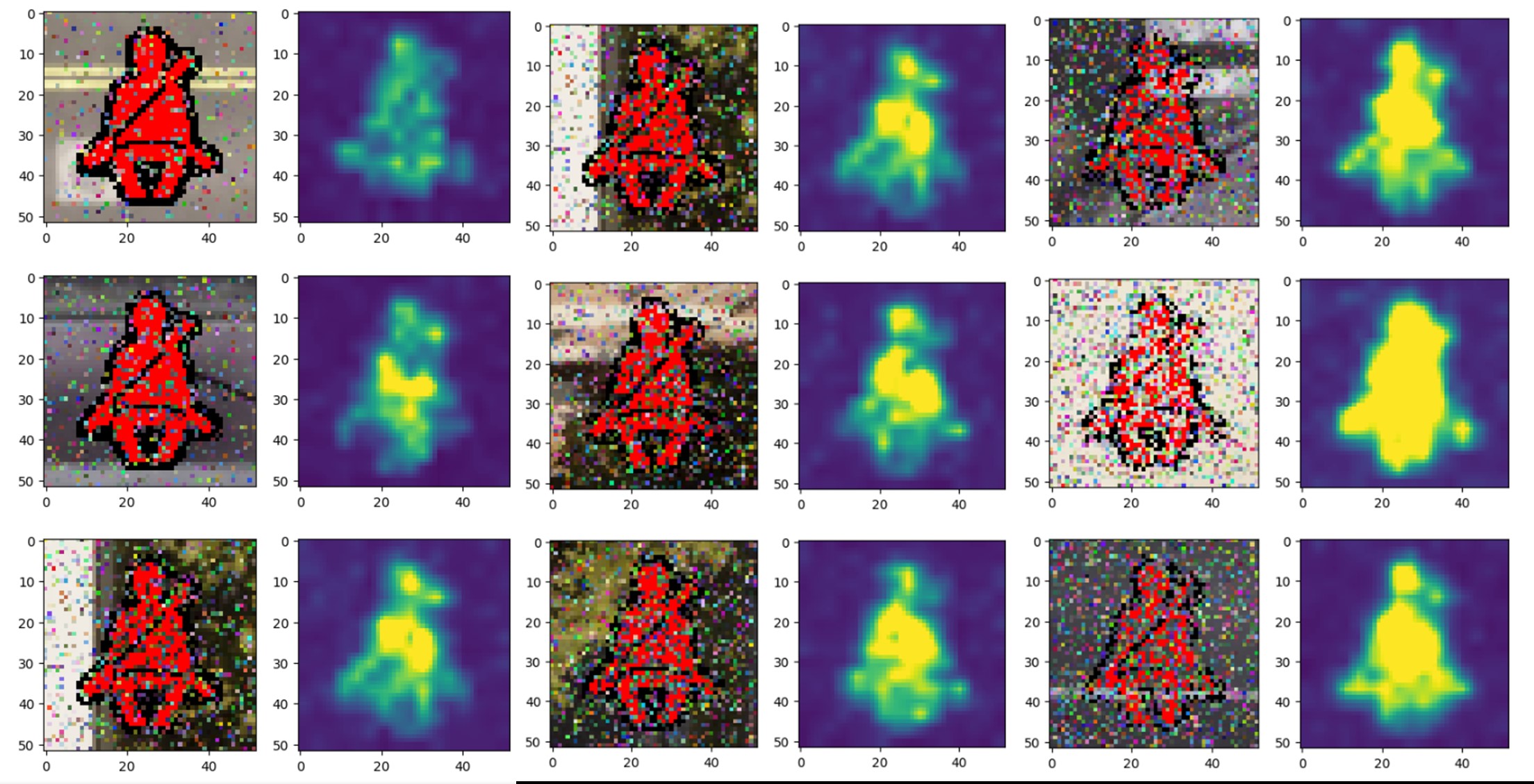}
         \caption{Noise error}
         \label{fig:subFigNoise}
     \end{subfigure}
        %\vspace{-0.1cm}
        \caption{Telltale degradation for different error types}
        \label{fig:Error_levels}
%        \vspace{-0.2cm}
    \vspace{-0.4cm}
\end{figure}

For the evaluation we injected a variety of different error types at different levels (magnitude), starting from small deviations, up to significant impact. The different types are depicted in Figure~\ref{fig:errors_examples}, while the influence of the error level is visualized in Figure~\ref{fig:Error_levels}. In detail, the following exemplary error types were used, which reflect the most relevant rendering errors according to~\cite{axmann2020supervising,bauer2019neue}:\begin{itemize}
    \item \textbf{No Render}: The telltale is not rendered at all.
    \item \textbf{Alpha Blending}: Telltale is blended over the background. With increasing error the transparency of the telltale is increased.
    \item \textbf{Color Error}: Error in the color of the telltale.
    \item \textbf{Pixel Noise}: Random modification of color value of the entire region.
     \item \textbf{Clipping Error}: Random erase of pixels on the telltale.
     \item \textbf{Partial Rendering}: Removal of parts of the telltale.
     \item \textbf{Stride}: Errors on the stride of the telltale during rendering.
     \item \textbf{Scale}: Increase of the scale of the telltale
\end{itemize} 

Most of the errors result in a degraded visibility, which increases with the magnitude of the error. E.g., a low level of \textbf{Pixel Noise} causes just a few erroneous pixels, which does not impact the readability of the telltale. In contrast a high error level impairs the visibility clearly. This is even more obvious for the \textbf{Alpha Blending}, where at the highest error level the telltale is fully transparent (Figure~\ref{fig:Error_levels}). Ideally we want the anomaly score to reflect this error quantity.  

In addition to the aforementioned errors, we also analyzed a more ``binary'' error type, where the image region is entirely filled with the dominant color of the telltale. The effect can be seen in Figure \ref{fig:errors_flood}. If the foreground is flooded, the telltale is not visible and the anomaly detection identifies that with a high anomaly score. If the background is flooded, the anomaly score remains low, yet the telltale also remains readable.

\begin{figure}[b!]
    \centering
    \vspace{-0.2cm}
    \includegraphics[width=0.95\columnwidth]{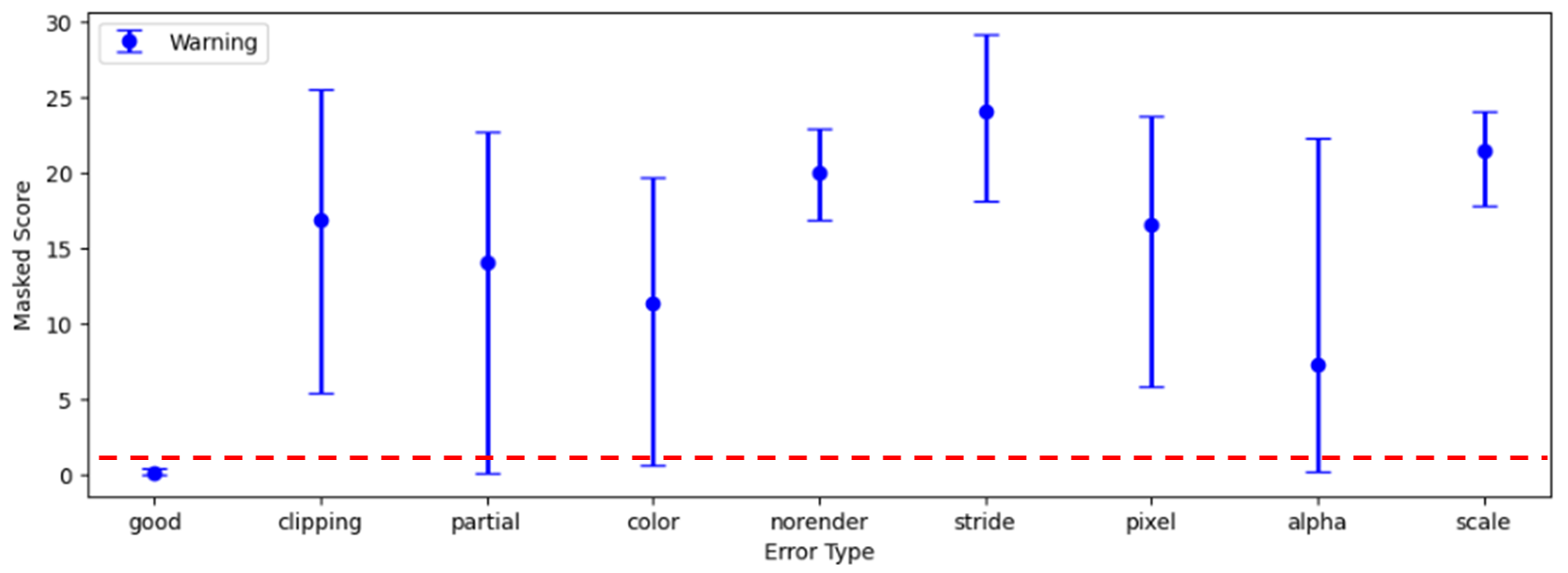}
    \caption{Distributions of the score for different error types (for the Warning telltale). The graphic shows the mean and the minimum and maximum values of the error type. Red line indicates separation among 'OK' and 'NOK'.}
    \label{fig:masterwarn_results}
\end{figure}

\subsection{Scoring for different Error Types \& Levels}

\begin{figure}[t!]
    \centering
    \includegraphics[width=0.95\columnwidth]{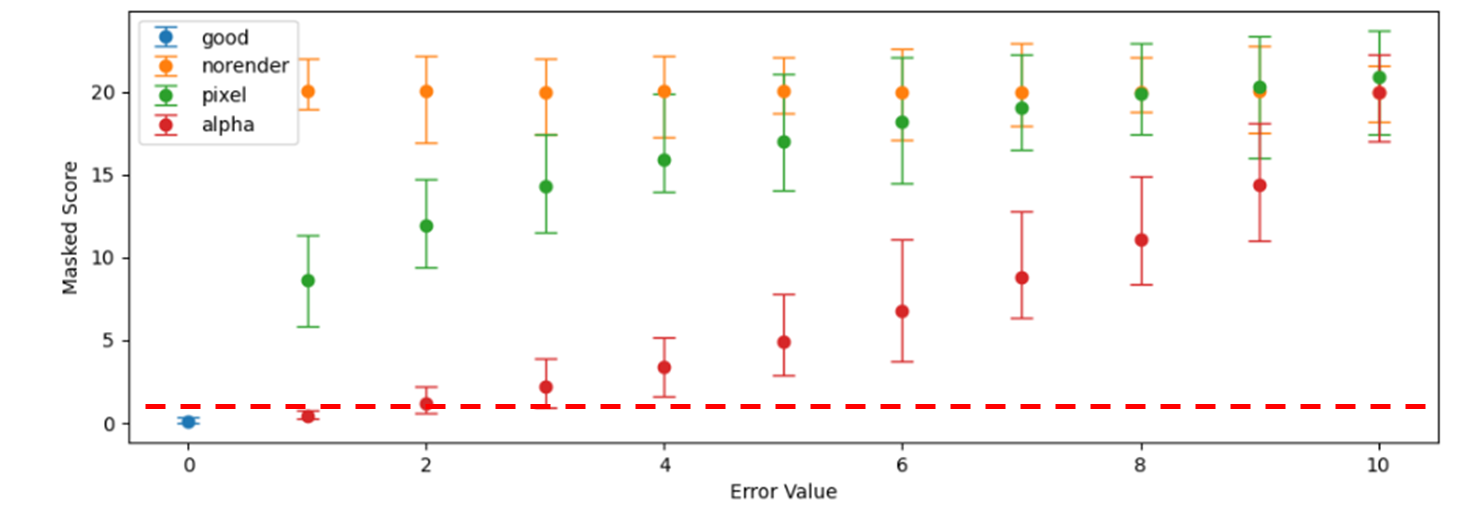}
    \caption{Distributions of the score for different error values / levels of alpha blending and pixel corruptions. The graphic shows the mean and the minimum and maximum values of the error type. Red line indicates separation among 'OK' and 'NOK'. 'norender' error doesn't change with the error levels, therefore mean is at same level for all error rates. }
    \label{fig:masterwarn_results_alpha}
    \vspace{-0.4cm}
\end{figure}

We evaluated the score for the different error types for the individual telltales. In Figure~\ref{fig:masterwarn_results} it can be seen that the good samples have a very low score and small variance. The different error types differentiate in the mean value, but for all error types the mean is significantly higher then the score of the good samples, which is a good indication that our solution can deliver the expected behavior. Furthermore, it can be noted that most errors show a high variance. This is due to the fact, that the score increases with the error levels (Figure~\ref{fig:masterwarn_results_alpha}). For low error levels the score is close to the good samples, which is also inline with the expectation as low error levels do not degrade the visibility noticeably. However, with an increase in error magnitude also the score increases, such that visible errors can be observed through a high score.

\begin{figure}[b!]    
    \centering
    \vspace{-0.6cm}
    \includegraphics[width=0.95\columnwidth]{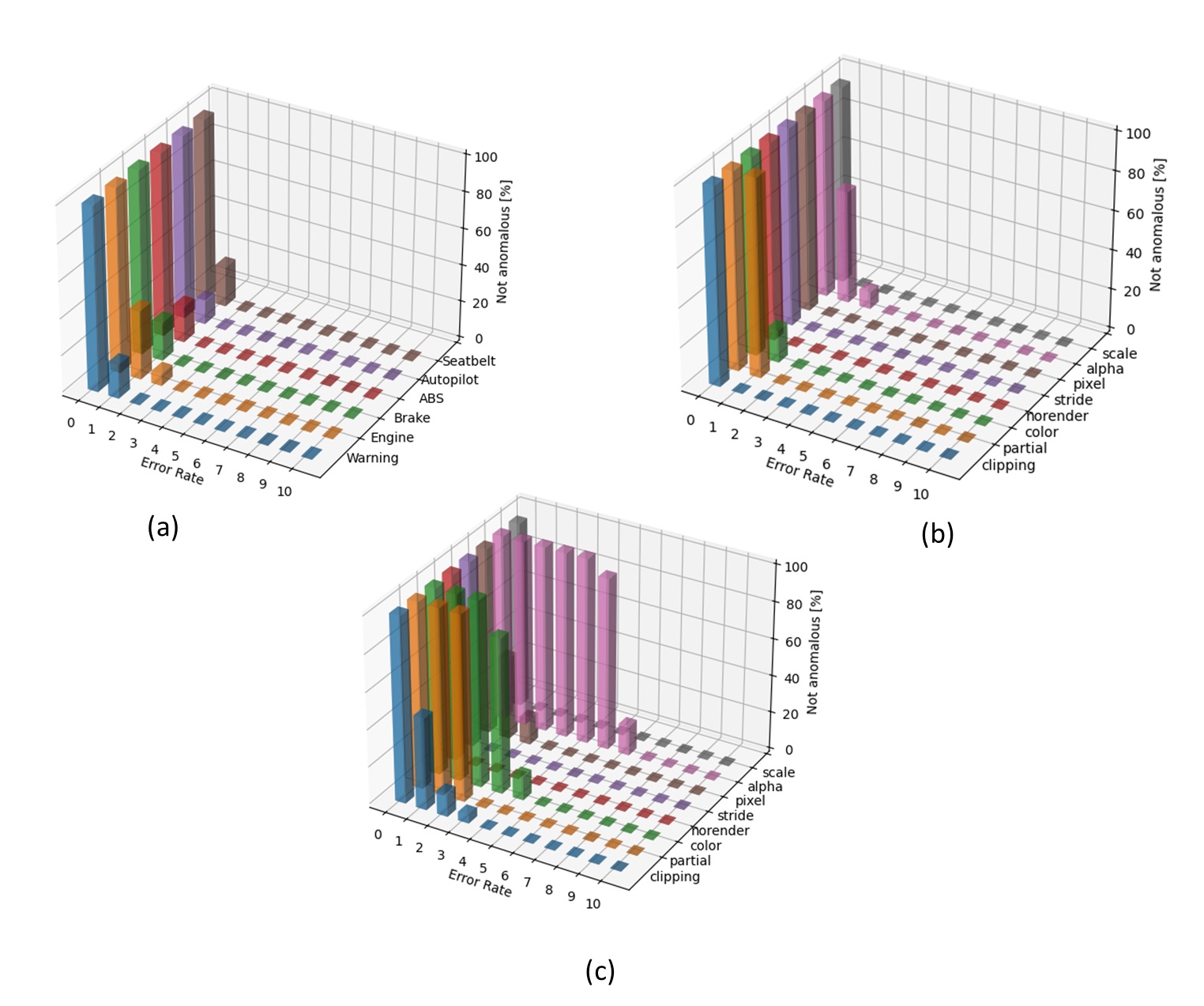}
    \caption{Final classification results using $\tau$ per telltale (a) or error type (b). (c) Results per error type with less rigorous threshold $\tau_{Alpha}$ }
    \label{fig:error_thresh}
\end{figure}

\begin{figure}[t!]
    \centering
    \includegraphics[clip,trim=.0cm .0cm 12cm 15.cm,width=0.95\columnwidth]{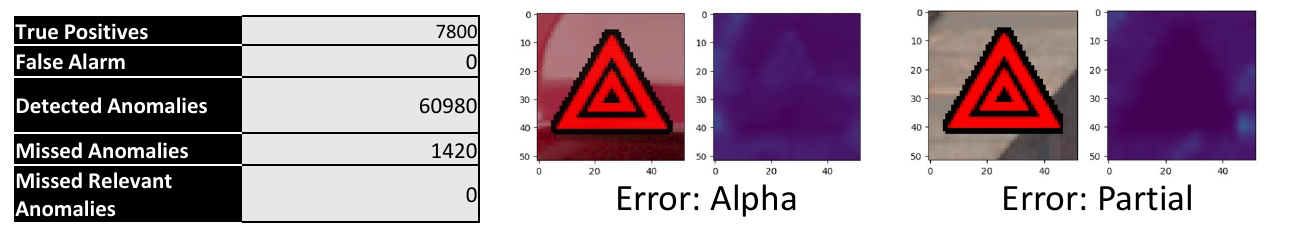}
    \caption{Detailed results using  $\protect\tau_{anomalous}^{Good}$. The not identified samples of error alpha and partial can be clearly identified by a human.}
    \label{fig:error_table}
    \vspace{-0.3cm}
\end{figure}

Figure~\ref{fig:masterwarn_results} and Figure~\ref{fig:masterwarn_results_alpha} also illustrate the threshold $\tau$ among 'OK' and 'NOK' telltales, which was obtained on the test dataset. To determine $\tau$, we use the following approach:
\begin{equation}
\tau = max_{score} (S_{test}^{Good}) \times m
\label{eq:thresh}
\end{equation}
In other words, the maximum score observable on the perfect samples of the test dataset ($S_{test}^{Good}$) is multiplied with a margin $m$ to account for variations in the scores given the higher variety of data in the final deployment. In our experiments we use $m = 2.1$, which showed a good balance between robustness and detection accuracy. With this configuration, rendering errors are robustly detected, expect very minor deviations in alpha or partial rendering, as show in Figure~\ref{fig:error_thresh} (a) and (b). As further detailed in Figure~\ref{fig:error_table}, this setting results in no false alarms, and only 1420 degraded telltales were not properly detected as defects. However, as these contain only minor errors that do not affect the understanding of the telltale or its visibility, all of these can be classified as not-relevant. Hence, the system is able in this setting to classify all error-free telltales correctly (no false alarms) and also can detect all relevant errors.

An advantage of our telltale monitor is the ability to handle alpha blending and other rendering effects. While this is already possible with the aforementioned configuration of $m = 2.1$, it may be desirable to allow even further reduced alpha settings on purpose. Therefore, one can define an additional
\begin{equation}
\tau_{alpha} = max_{score} (S_{test}^{Alpha}[error\_level]) \times m
\label{eq:thresh_alpha}
\end{equation}
As illustrated in Figure~\ref{fig:error_thresh}(c), this setting can still identify all important rendering errors, but is more relaxed for alpha blending, color drifts and partially missing telltale elements.

With the same approach we can as well accomplish checks for warping and scaling of telltales. To achieve that we de-warp / de-scale the transformed telltale back to its original shape. Usually this will create some artifacts in the resulting image. But with the proven tolerance to minor deviations we can still separate correct from anomalous samples.

\subsection{Scoring for different telltales}

\begin{figure}[b!]
    \centering
    \vspace{-0.2cm}
    \includegraphics[clip,trim=0.cm .0cm 0cm 0.cm,width=0.98\columnwidth]{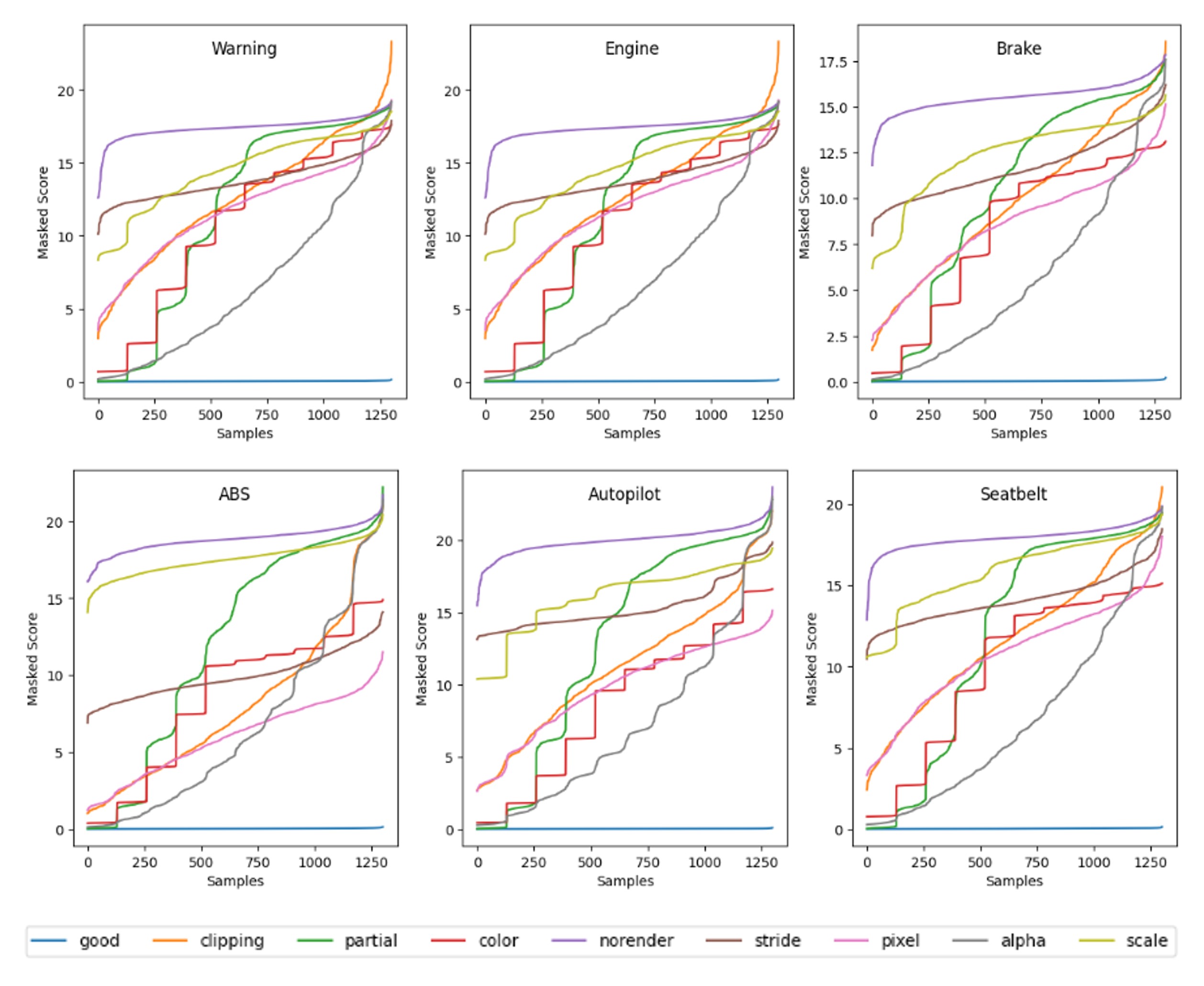}
    \caption{Overview of scores for all evaluated telltales. The individual scores are sorted in ascending order per error type.}
    \label{fig:scores_all}
\end{figure}

The previous discussions have been focused on the results of a single telltale. However, it is important that the same results can be achieved for the other telltales as well. Figure~\ref{fig:scores_all} shows the scores for all the six telltales we used in our evaluation. It can be seen that the results slightly vary in absolute numbers, but that the overall behavior is comparable. The scores of the good samples is always close to zero, and the injected rendering errors have in all cases a similar degrading effect. This shows that our solution can address different telltales with different shape or color.

\subsection{PCA Decomposition}

\begin{figure}[t!]
     \centering
     \begin{subfigure}[b]{0.8\columnwidth}
     \vspace{-0.1cm}
         \centering
         \includegraphics[clip,trim=0.0cm 0.cm 0.0cm 0.cm,width=\textwidth]{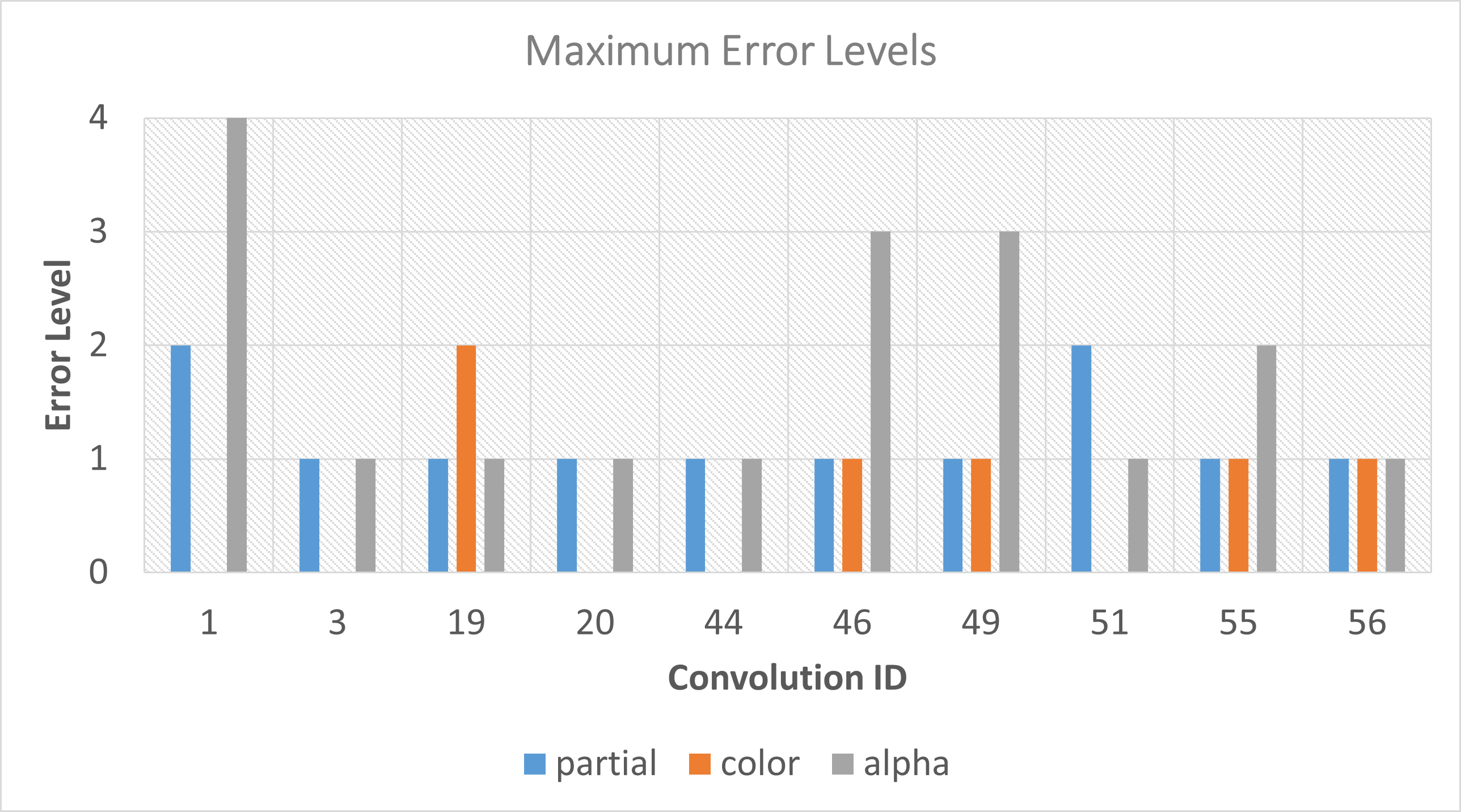}
         \vspace{-0.5cm}
         \caption{Max error level that is 'OK'}
         \label{fig:multiplePCAs}
     \end{subfigure}
     \newline
      \begin{subfigure}[b]{0.95\columnwidth}
          \centering
          \includegraphics[clip,trim=0.0cm 0.cm 0.0cm 0.cm,width=\textwidth]{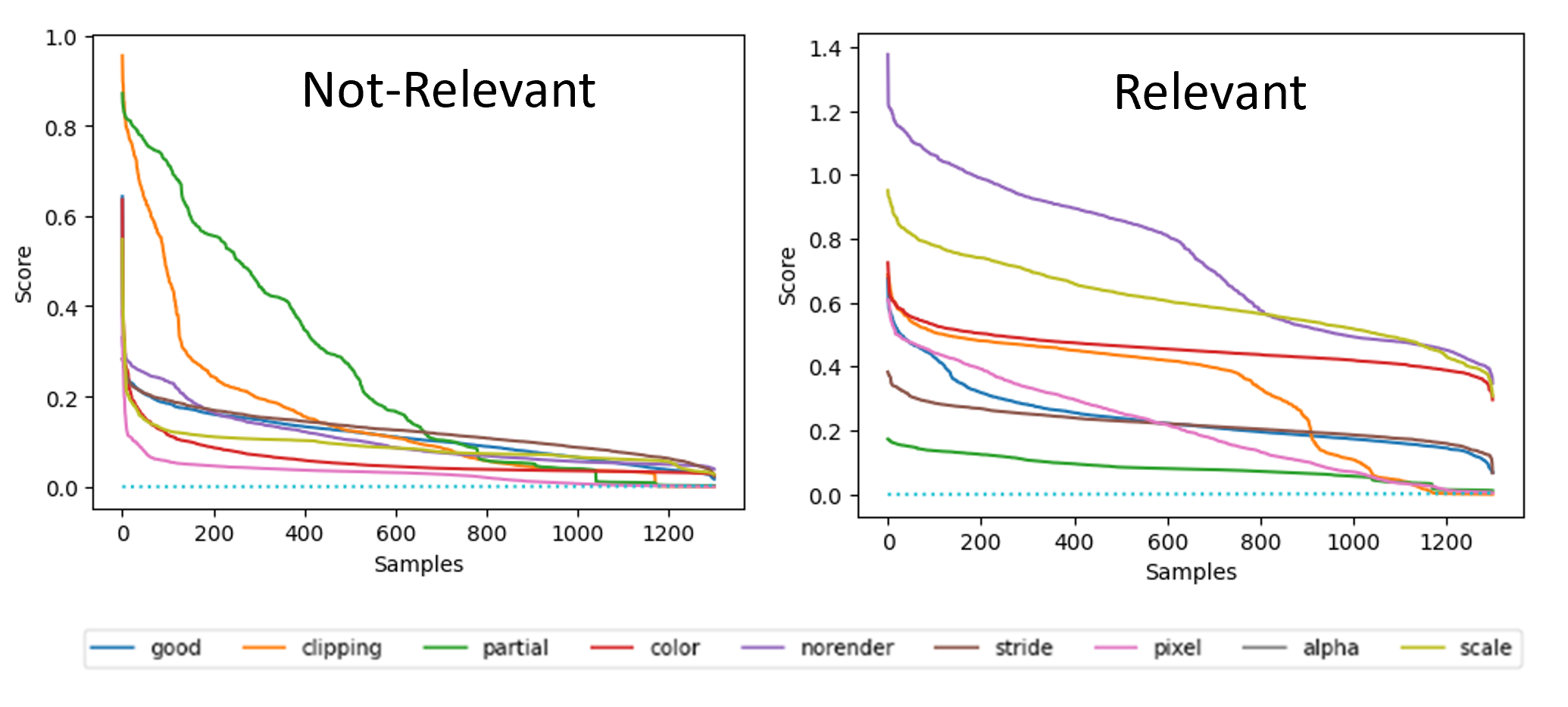}
          \vspace{-0.5cm}
          \caption{Two different example PCAs}
          \label{fig:SignlePCA}
      \end{subfigure}
         \vspace{-0.1cm}
         \caption{Influence of individual PCAs on score and error segmentation}
%         \label{fig:Error_levels}
% %        \vspace{-0.2cm}
     \vspace{-0.4cm}
\end{figure}

For now we used all features for training and scoring with a single PCA, i.e. the entire feature tensor was fed into a single PCA. As motivated in Section~\ref{sec:solution}, using a single large PCA may not be the best choice. Therefore, we analyzed the contribution of the individual features (convolutions) to the final scoring. To achieve this, 64 PCAs were trained, each of them on just one feature. For each of the PCAs we determined $\tau_{anomalous}^{Good}$ as previously described. 
With $m=1.2$, only 10 of 64 PCAs showed a meaningful influence on the final score, and classified the good samples correctly. For the remaining PCAs, no clear separation is possible, as illustrated in Figure~\ref{fig:SignlePCA}. For one of the relevant PCAs, Figure~\ref{fig:multiplePCAs} depicts the classification results, showing that the maximum tolerated error level for this PCA is 4, which still is acceptable as humans can perceive such telltales correctly. Yet, there are also PCAs that have an even stricter separation and tolerate only lower levels. This shows that in fact only a subset of features can be used within the safety monitor, and that only the most relevant features should be used to achieve the most efficient realization (up to 10x faster). However, it is worth noting that not just a single PCA on a single feature should be used, as this could cause a higher undetectable error rate, which is address in the next section.

\begin{figure*}[t!]
    \centering
    \includegraphics[width=0.85\textwidth]{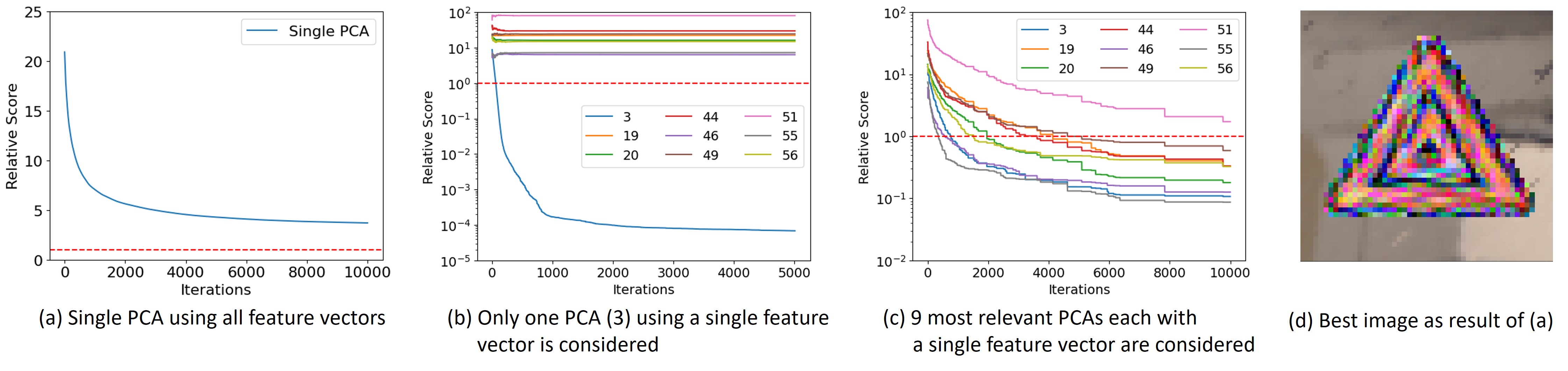}
    %\vspace{-0.2cm}
    \caption{Generation of an ``undetectable error'' using different PCA configurations. Relative score is score over $\tau$. Dotted line separates ’OK’ and ’NOK’ scores.}
    \label{fig:attacks}
    \vspace{-0.45cm}
\end{figure*}

\subsection{Safety Discussion \& Undetectable Errors}
\vspace{-0.1cm}

A great challenge of learning-based safety monitors is to provide evidence that the entire system (main function + safety monitor) is safe enough. For a CRC-based safety monitor it is possible to mathematically prove that it can identify all single pixel errors, but this aspect also limits the overall system functionality (e.g. overlays are hardly possible as explained in Section~\ref{sec:background}). For our proposed monitor, the roles are inverted, as the desired system functionality can be achieved, but the safety evidence is harder to achieve. Therefore, some key considerations are provided next.

As mentioned in Section~\ref{sec:solution}, our safety monitor covers both cases, errors for active as well as inactive telltales. However, while a rendering error for an inactive telltale is certainly inconvenient, only an error for an active telltale is safety-relevant. Hence, for a critical error the related telltale needs to be active and a rendering error has to occur. As discussed before, many rendering errors are properly detected, however, there is a certain chance for undetectable errors, meaning corrupted input images that are not properly flagged by the safety monitor as 'NOK'. 

To test if our safety monitor is susceptible to such images, we used a genetic algorithm (GA) to find such input images. The goal of the GA is to find an image with a score sufficiently low to be acceptable, i.e. the safety monitor would consider the image as a valid telltale. As initial population, 1000 images were selected, which contained up to 5\,\% faulty pixels within the telltale shape for a given background. With each new generation, the amount of faulty pixels can increase through the manipulation and crossover, providing a wide variety of images that are tested.

As illustrated in Figure~\ref{fig:attacks}, there is a significant difference in undetectable errors, depending on the PCA configuration. If only a single PCA is used covering all features, no undetectable error was found within $10^4$ generations (each covering a population of 1000 images). Instead, if PCAs are used that address only a single feature (see Section 4.4), undetectable errors can be found. In this case, the overall undetectable error rate depends on how many PCA scores are combined to create the final score. If all of them need to be 'OK', as shown by Figure~\ref{fig:attacks}(c), no undetectable error was found in $10^4$ generations, while if a single PCA score is sufficient, less than 100 generations are required (Figure~\ref{fig:attacks}(b)). Hence, depending on the safety requirement of the telltale, a different setting can be chosen that achieves the required safety level with best performance and compute demands (fewer single-feature PCAs is faster). For example, an informational telltale (e.g. high beam on/off), can use only the most relevant PCAs, whereas a critical telltale (e.g. brake failure) can use a PCA that contains all possible features.

In any case, it is worth noting that the results also show that there are many more detectable than undetectable errors. In addition, it is worth noting that the image that causes the safety monitor to fail, is a combination of random pixel values, which can be clearly detected by other means. 
Furthermore, as depicted in Figure~\ref{fig:attacks}(d), the obtained image that got closest to an undetectable error, shows already the black edges of the correct telltale. In fact, the more iterations were executed, the more did the created image converge towards the correct telltale. This underlines that the safety monitor ``learned'' relevant telltale features.

\begin{figure}[b!]
     \centering
     \vspace{-0.4cm}
     \includegraphics[width=0.9\columnwidth]{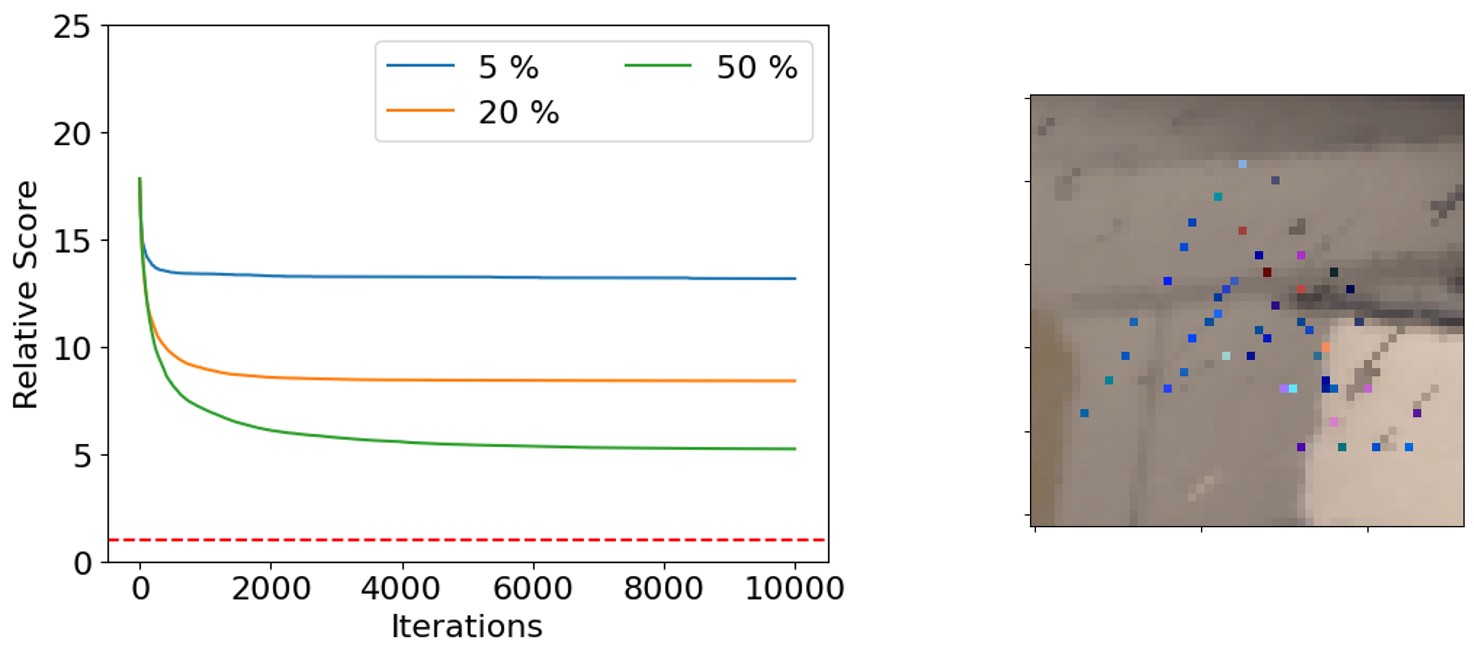}
     \caption{Score evolution with a restricted number of corrupted pixels for the same experiment as in Figure~\ref{fig:attacks}. Relative score is score over $\tau$. Dotted line separates ’OK’ and ’NOK’ scores.}
     \label{fig:attack_restricted}     
\end{figure}

Furthermore, we performed tests, where we intentionally limited the number of corrupted pixels. As a result, all images show basically an empty background with only a few faulty pixels (see Figure~\ref{fig:attack_restricted}). If the safety monitor would flag one of these images as valid, this may result in a safety relevant error, as a human driver would not be able to understand that a telltale should be displayed. To test these situations, we used the same GA as before, but limited the number of corrupted pixels after each generation to at most 5\,\%, at most 20\,\% and at most 50\,\%. As it can be inferred from Figure~\ref{fig:attack_restricted}, under these conditions the GA was not able to generate an image with a sufficiently low score, i.e. all generated and tested images were correctly flagged as containing no valid telltale. It is also clearly visible that fewer pixel errors lead to higher scores, which is inline with the expectation. Consequently, it can be concluded that just a few random pixel faults on an empty background will not cause an undetected error. In fact a huge amount of pixel errors needs to be present to make the safety monitor flag a rendered image as being a valid tell-tale, which can then be easily detected by the driver.

\section{Conclusion}
\label{sec:conclusion}
The expectations for the capabilities of digital instrument cluster are increasing with each new automotive generation. Classic dials are now a thing of the past, while instead, users desire beautiful 3D navigation maps, design customization options, and smartphone-like rendering, including modern effects such as overlays or color grading. This, however, imposes a challenge for the state-of-the-art error detection approaches used to detect rendering errors. Therefore, we presented in this work a novel learning-based error detector evaluated for the rendering of telltales. The monitor utilizes a feature-based principal component analysis to detect rendering errors that would disturb the human perception of a telltale. Our evaluation results show that the system is able to detect all significant errors, while not causing false alarms on correctly rendered telltales.

\bibliographystyle{IEEEtran}
\bibliography{main}

\end{document}